\pdfoutput=1

\documentclass[11pt]{article}
\usepackage[final]{acl}
\usepackage{xcolor}
\definecolor{customRed}{HTML}{C04F15}
\definecolor{customGreen}{HTML}{3B7D23}
\definecolor{customLightRed}{HTML}{FBE3D6}
\definecolor{customBlue}{HTML}{215F9A}

\usepackage{tabularx}
\usepackage{times}
\usepackage{latexsym}
\usepackage{booktabs}
\usepackage{multirow}
\usepackage[table,xcdraw]{xcolor}
\definecolor{lightblue}{HTML}{F1F9FF}
\definecolor{lightgreen}{HTML}{f6fbf4}
\definecolor{lightpink}{HTML}{FFF3F4}
\usepackage[T1]{fontenc}
\usepackage{float}
\usepackage[linesnumbered,ruled,vlined]{algorithm2e}
\usepackage{xspace}
\SetKwComment{Comment}{$\triangleright$\ }{}
\SetAlCapNameFnt{\normalfont}
\SetAlCapFnt{\small}
\SetKwInOut{Input}{\textbf{Input}}
\SetKwInOut{Output}{\textbf{Output}}
\usepackage{amsmath}
\usepackage{amsmath, amssymb}

\usepackage[utf8]{inputenc}
\usepackage[justification=raggedright,singlelinecheck=false]{caption}
\SetKwComment{Comment}{$\triangleright$\ }{}
\usepackage{microtype}
\usepackage{amsmath} 
\usepackage{inconsolata}
\usepackage{cleveref}
\usepackage{graphicx}
\usepackage{enumitem}  

\title{From Query to Counsel: Structured Reasoning with a Multi-Agent Framework and Dataset for Legal Consultation}

\author{
  Mingfei Lu\textsuperscript{1} \quad
  Yi Zhang\textsuperscript{1}\thanks{\ Corresponding authors.} \quad
  Mengjia Wu\textsuperscript{1} \quad
  Yue Feng\textsuperscript{2}\footnotemark[\value{footnote}] \\
  \textsuperscript{1}Australian Artificial Intelligence Institute (AAII),
  University of Technology Sydney \\
  \textsuperscript{2}School of Computer Science, University of Birmingham \\
  \texttt{mingfei.lu@student.uts.edu.au} \\
  \texttt{\{yi.zhang, mengjia.wu\}@uts.edu.au},
  \texttt{y.feng.6@bham.ac.uk}
}

\usepackage{algorithmic}
\begin{document}
\maketitle
\begin{abstract}
Legal consultation question answering (Legal CQA) presents unique challenges compared to traditional legal QA tasks, including the scarcity of high-quality training data, complex task composition, and strong contextual dependencies. To address these, we construct \textsc{JurisCQAD}, a large-scale dataset of over 43,000 real-world Chinese legal queries annotated with expert-validated positive and negative responses, and design a structured task decomposition that converts each query into a legal element graph integrating entities, events, intents, and legal issues. We further propose \textsc{JurisMA}, a modular multi-agent framework supporting dynamic routing, statutory grounding, and stylistic optimization. Combined with the element graph, the framework enables strong context-aware reasoning, effectively capturing dependencies across legal facts, norms, and procedural logic. Trained on \textsc{JurisCQAD} and evaluated on a refined LawBench, our system significantly outperforms both general-purpose and legal-domain LLMs across multiple lexical and semantic metrics, demonstrating the benefits of interpretable decomposition and modular collaboration in Legal CQA.

\end{abstract}

\section{Introduction}

\begin{figure}[h]
  \raggedright
  \includegraphics[width=0.50\textwidth]{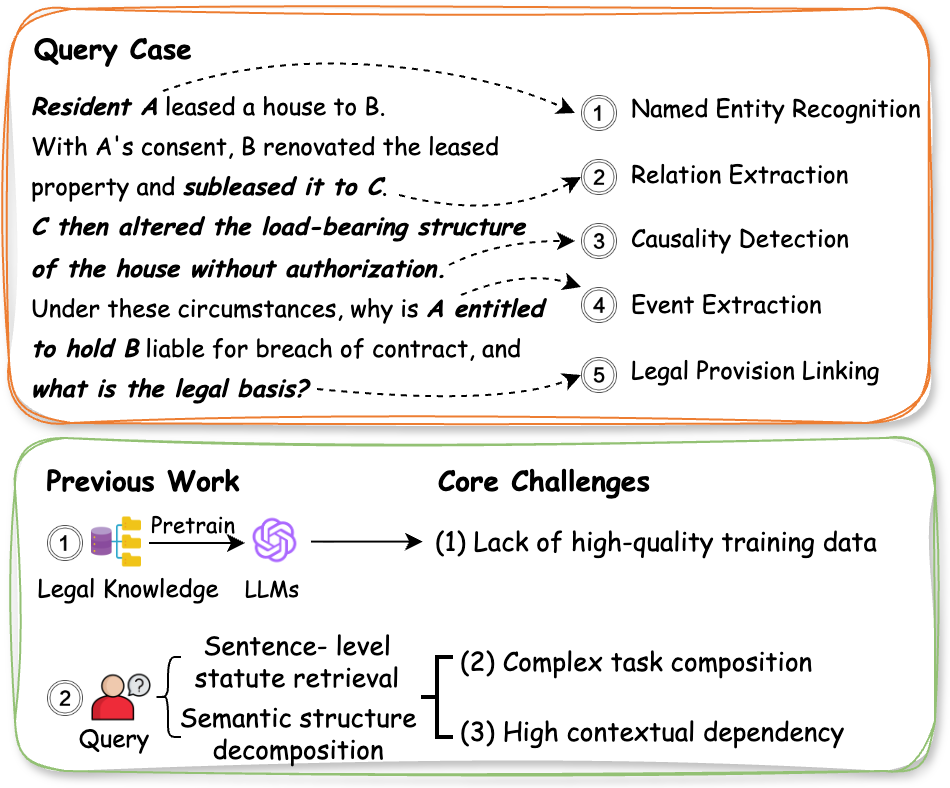}
\caption{
An illustrative example of legal consultation task decomposition, highlighting key challenges, limitations of prior approaches}
\label{Figure1}
\end{figure}

Legal consultation question answering (Legal CQA) is an emerging core task in legal artificial intelligence. It targets real-world user queries involving personalized legal dilemmas, and aims to generate contextually grounded, executable legal advice~\citep{zhong-etal-2020-nlp,louis2024interpretable}. Its practical value, social impact, and high complexity make it an essential benchmark for evaluating legal reasoning capabilities in language models.

Consider the example in Figure~\ref{Figure1}.
Answering such a query requires:
(1) Identifying legal relationships - recognizing A and C lack direct privity while B serves as the contractual intermediary;
(2) determining causation - tracing the damage to C's unauthorized structural alteration;
(3) identifying the core issue - explaining why A can bypass C to hold B liable;
(4) reasoning and legal matching - based on the above analysis, concluding that the principle of contractual relativity and B's supervisory obligations apply, then matching with corresponding civil code provisions.
This example illustrates that Legal CQA demands: (1) Strong legal background knowledge, (2) legal-oriented query decomposition, and (3) powerful contextual understanding and reasoning.

Previous studies on Legal CQA generally follow two main workflows. The first line of work seeks to enhance legal knowledge within large language models by continuing pretraining on legal statutes, judicial opinions, and domain-specific corpora~\citep{huang2023lawyerllamatechnicalreport}. This strategy aims to enrich the model’s understanding of legal terminology, doctrinal structures, and statutory patterns in an unsupervised manner—but due to coarse data processing, limited coverage, and non-scenario-based sources, the resulting supervision is low-quality, leading to marginal performance gains. The second line of work focuses on input restructuring, such as retrieving sentence-level legal provisions to provide explicit legal grounding for generation~\citep{ma2023caseencoder,ni2025pre}. These methods typically rely on retrieval pipelines to extract relevant statutes or past cases, which are then concatenated with the user query to inform the model’s response. However, these methods fall short in handling real-world legal consultations, which are often vague, multi-faceted, and require dynamic interpretation of facts, actors, and legal implications beyond static law matching.

We thus identify three core challenges in Legal CQA: 1) \textbf{Lack of high-quality training data} that reflects realistic legal consultation scenarios; 2) \textbf{complex task composition}, involving multiple, interdependent subtasks; and 3) \textbf{high contextual dependency}, requiring precise interpretation of legal entities, relationships, and user intent.

To address these challenges, we present \textsc{JurisCQAD}, a large-scale benchmark dataset comprising over 43,000 real-world legal consultation instances, each organized as a triplet of (question, positive/negative answer). This dataset supports open-ended, generative Legal CQA tasks and enables high-quality model training and evaluation.
Building on \textsc{JurisCQAD}, we propose \textsc{JurisMA}, a multi-agent framework designed to simulate real-world legal decision-making. \textsc{JurisMA} decomposes complex consultation tasks into structured graph-based elements, capturing legal entities, relationships, intents, and issues. Then it employs a cooperative multi-agent architecture to solve these subtasks and reach consensus. The method is evaluated against general large language models (LLMs) and legal-specialized LLMs, achieving state-of-the-art performance across multiple evaluation metrics via training on \textsc{JurisCQAD}. 

\begin{figure*}[t]
  \centering
  \includegraphics[width=\textwidth]{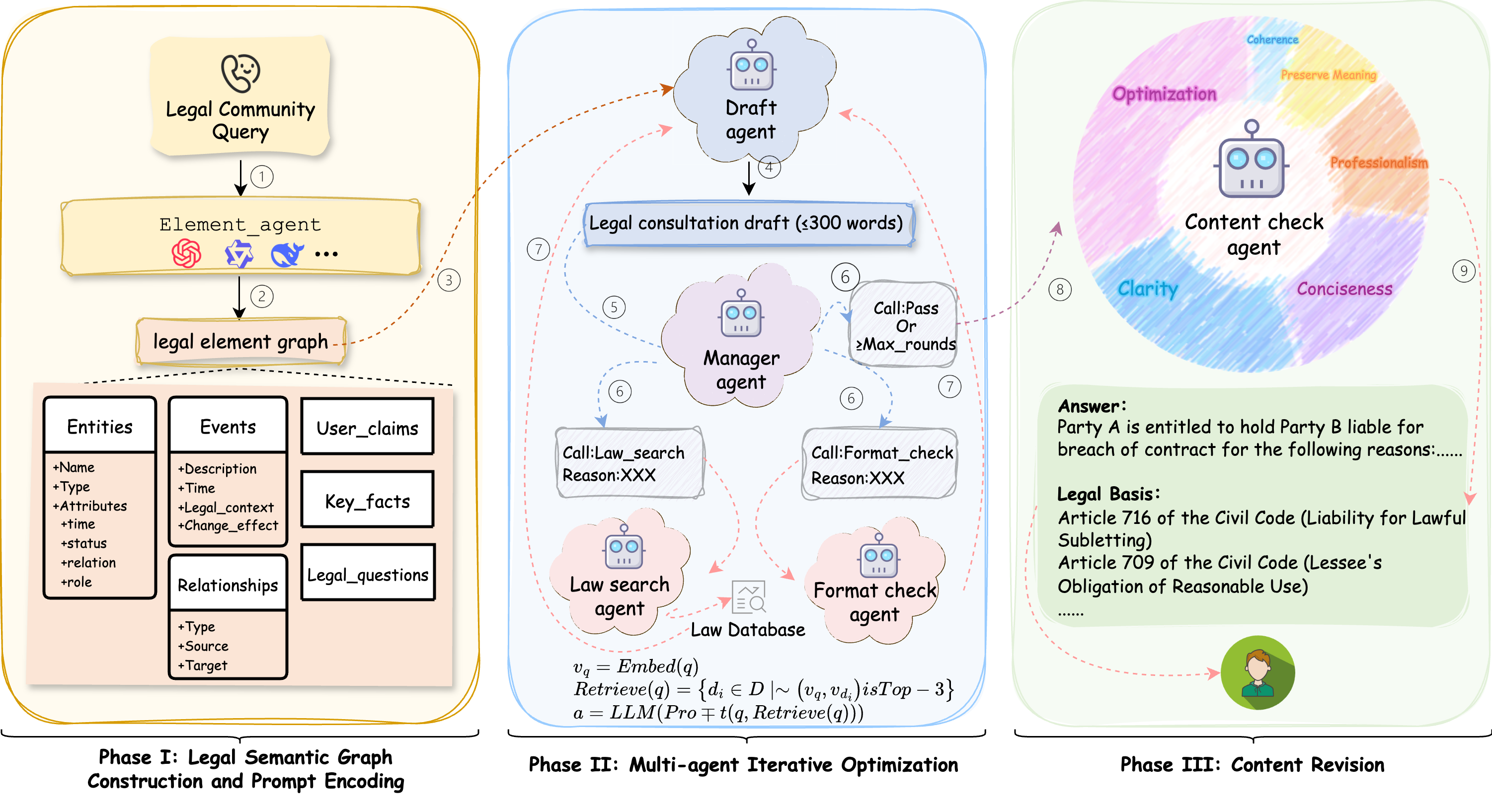}
\caption{
Overview of \textsc{JurisMA}, a multi-agent framework that parses legal queries into element graphs, refines drafts via agent collaboration, and outputs a final legal opinion with supporting statutes.}
  \label{Figure2}
\end{figure*}

The contribution of this work is three-fold:

(1) A large-scale, high-quality dataset, tailored for Legal CQA: It enables model training and evaluation, with significant improvement in generating accurate and context-aware legal responses.

(2) A structured graph-based task decomposition strategy: It extracts legal entities, events, relationships, user intents, and legal issues from free-form queries, offering a flexible and effective framework for Legal CQA.

(3) A pluggable and modular multi-agent system where a Manager Agent dynamically coordinates subtasks across specialized agents through multi-round refinement: This framework demonstrates strong performance and adaptability in handling high-context legal consultations.

\section{Related Work}
\textbf{Evolution of Legal QA Methodologies.}

Early systems relied on traditional retrieval methods such as BM25~\citep{shao2020bert,jayawardena2024scale}, which performed well on structured statute lookup but struggled with ambiguity and long-form queries. With the rise of generative LLMs, domain-adapted models such as 
LawGPT~\citep{zhou2024lawgptchineselegalknowledgeenhanced} have enhanced semantic understanding via pretraining, yet suffered from uncontrollable 
reasoning and legal inconsistency, motivating broader efforts on 
adapting, refining, and stabilizing LLMs for downstream 
use~\citep{lu2025tpa,yangwalking,caisteering,lu2026choosing,lu2025newborn}. In contrast, our multi-agent framework leverages a Manager Agent to coordinate subtasks and synchronize legal basis updates during iterative review, improving both completeness and validity.

\textbf{Legal Knowledge Representation and Augmentation.}
Static injection methods (e.g., LEGAL-BERT~\citep{shao2020bert}) enrich legal embeddings but struggle with evolving laws. Retrieval-augmented methods (e.g., LSIM~\citep{yao2025elevating}) offer real-time updates via semantic similarity, but often confuse legally distinct yet linguistically similar terms. Our structured semantic graph models entity–relation–fact chains, reducing ambiguity and enhancing interpretability.

\textbf{Multi-Agent Approaches in Legal Tasks.}
Existing legal multi-agent systems often follow rigid pipelines (e.g., LawLuo~\citep{zhang2025survey}), limiting adaptability in real consultations. General frameworks like ReAct~\citep{yao2023react} support dynamic reasoning, but lack legal-domain compliance checks. We introduce a Manager Agent that dynamically assesses draft quality and coordinates cooperative repair via FormatCheck and LawSearch agents, mitigating error propagation and enhancing legal robustness.

\textbf{Data Resources and Evaluation.}
Datasets like LegalQA~\citep{nigam2023legal} and LLeQA~\citep{louis2024interpretable} focus on statute retrieval or synthetic Q\&A, lacking realism and linguistic diversity. This problem is acute in Chinese legal NLP, where most models (e.g., LawGPT) are trained on artificial data, leading to domain shift. We address this gap by constructing a 43K-scale dataset of real-world legal consultations with expert-verified triplet annotations, covering high-frequency domains and supporting robust, grounded evaluation.

\section{Methodology}

In this section, we introduce the task formulation, followed by a detailed description of \textsc{JurisMA}. We then describe the test dataset correction procedure, the construction of \textsc{JurisCQAD}, and the model training process.

\subsection{Task Formulation}

Legal CQA differs fundamentally from traditional legal QA tasks—such as statute retrieval or multiple-choice assessments—which operate on well-defined inputs and constrained options. In contrast, Legal CQA deals with open-domain user queries expressed in natural language, often characterized by factual ambiguity, informal phrasing, and personalized legal dilemmas. These queries commonly involve multiple entangled legal concepts and reasoning steps. A detailed comparison is provided in Table~\ref{tab:qa_comparison}.


Formally, given a user query \( q \in \mathcal{Q} \), the objective of Legal CQA is to generate a response \( r \in \mathcal{R} \) that satisfies the following criteria: (1) \(\mathrm{Align}(r, q)\): Semantic alignment with the user's factual context and legal intent; (2) \(\mathrm{Legal}(r)\): Compliance with applicable Chinese legal provisions; (3) \(\mathrm{Express}(r)\): Clear, accurate, and professional expression.

Most existing systems use an end-to-end generation paradigm. While effective in general NLP tasks, this approach often ignores the structured and step-by-step nature of legal reasoning.
To address this gap, we propose a modular and interpretable pipeline inspired by cognitive theories of legal writing.
Our method breaks the task into three sequential phases, as shown in Figure~\ref{Figure2}.
These phases share a common semantic context—captured by an element graph—and are managed by a centralized controller. This design supports iterative refinement and controllable reasoning.

\subsection{Phase I: Legal Semantic Graph Construction and Prompt Encoding}

This phase transforms user queries into structured legal element graphs that provide a semantic foundation for downstream reasoning. Legal reasoning revolves around identifying key facts, involved entities, and legal relationships. 

Motivated by this observation, we design an \textit{Element Agent} to construct a graph-based representation \( G \) that explicitly encodes legal entities, events, and their semantic connections.

This design is grounded in jurisprudential theories that treat law as a structured system of subjects, facts, norms, and relationships. It aligns with Hart’s distinction between primary and secondary rules~\citep{hart2012concept}, and Kelsen’s hierarchical model of normative systems~\citep{kelsen1967pure}.

As illustrated in Figure~\ref{Figure2}, we define the legal element graph as \( G = (V, E) \), where \( V \) is the set of nodes and \( E \) the set of edges. The node set \( V \) includes: (1) Entities: individuals or organizations (e.g., plaintiffs), annotated with attributes such as roles, statuses, and timestamps; (2) Events: legal actions or disputes; (3) User claims, key facts, and inferred legal questions.

Edges \( E \) represent semantic relations among elements, such as kinship and contractual duties.

The graph \( G \) is serialized in JSON and serves as a global contextual abstraction. This structure enhances both the interpretability and controllability of downstream reasoning modules. An example is provided in Appendix~\ref{sec:Example of Element Graph}.

To prepare the initial generation, we serialize the element graph \( G \) into a prompt \( P_G \), and concatenate it with the user query \( q \) to form the input for generation:
\ $
u = [P_G ; q],
$ \
This structured input provides rich semantic grounding for downstream response generation.

\subsection{Phase II: Multi-agent Iterative Optimization}

\begin{algorithm}[t]
\caption{Multi-Agent Controlled Draft Optimization}
\label{alg:draft_optimization}
\textbf{Input}: User query $q$; element graph $G$; initial draft $r_0 = f([P_G; q])$\\
\textbf{Initialize}: $t \leftarrow 0$, $r_t \leftarrow r_0$\\
\textbf{Output}: Final legal response $r_{\text{final}}$
\begin{algorithmic}[1]
\WHILE{$t < T$}
  \STATE $a_t \leftarrow \text{ManagerAgent}(r_t)$
  \IF{$a_t = \emptyset$ \OR $a_t = \texttt{"Pass"}$}
    \STATE \textbf{break} \COMMENT{Stop if draft is acceptable}
  \ENDIF
  \IF{$\texttt{"FormatCheck"} \in a_t$}
    \STATE $s_t \leftarrow \text{FormatCheckAgent.getSuggestions}(r_t)$
    \STATE $r_t \leftarrow \text{DraftAgent.applySuggestions}(r_t, s_t)$
  \ENDIF
  \IF{$\texttt{"LawSearch"} \in a_t$}
    \STATE $l_t \leftarrow \text{LawSearchAgent.retrieveStatutes}(q, r_t)$
    \STATE $r_t \leftarrow \text{DraftAgent.integrateStatutes}(r_t, l_t)$
  \ENDIF
  \STATE $t \leftarrow t + 1$
\ENDWHILE
\STATE $r_{\text{final}} \leftarrow \text{ContentCheckAgent}(r_t)$
\STATE \textbf{return} $r_{\text{final}}$
\end{algorithmic}
\end{algorithm}

This phase focuses on refining the initially generated legal draft through a multi-agent framework that supports structured, iterative optimization.

Given the semantic input \( u = [P_G ; q] \) prepared in the previous phase, we obtain an initial response \( r_0 = f(u) \) using a \textit{Draft Agent}. This preliminary output is then subject to refinement via downstream agents coordinated by a centralized controller called the \textit{Manager Agent}. It follows the modular control paradigm in multi-agent systems, where a centralized planner selectively activates agents based on intermediate output quality~\citep{russell2016artificial}. 

The iterative optimization process emulates professional legal writing workflows, where drafts are collaboratively refined by experts with distinct roles. In our system, \textit{Manager Agent} oversees the process by dynamically assessing whether the current draft requires further improvement. Its decisions are guided by two primary criteria: (1) linguistic adequacy, including clarity and conciseness; and (2) legal completeness, particularly the inclusion of authoritative statutory references.

If deficiencies in structure or expression are detected, the \textit{Format Check Agent} is invoked to generate targeted revision suggestions. These suggestions are then integrated by the \textit{Draft Agent} in the subsequent iteration. If legal references are missing or insufficient, the \textit{Law Search Agent} will retrieve relevant provisions from statutory databases.

This multi-agent refinement loop proceeds for up to five iterations or until the \textit{Manager Agent} returns a “Pass” signal. The modular design ensures that only necessary sub-agents are activated, maintaining both efficiency and controllability. 

The refinement leverages an iterative feedback loop similar to recent approaches in planning-based generation and multi-pass text optimization~\citep{zhang2020pegasus}. This design allows for gradual correction and quality enhancement until the response meets legal, factual, and stylistic requirements.The full procedure is shown in Algorithm~\ref{alg:draft_optimization}.

\subsection{Phase III: Content Revision}
This phase aims to refine the draft into a coherent, professional, and legally structured output.

After all iterative revisions, a \textit{Content check Agent} is responsible for rewriting the legal opinion to meet professional norms. The agent receives the original user query and the revised draft as input. It then performs a final pass focused on language quality and output structuring, without introducing new legal content or altering legal positions.

Specifically, the \textit{Content Check Agent} preserves the original legal meaning, improves clarity, precision, and conciseness, and outputs the result in a dual-section format.

The finalized output is partitioned as follows:
\begin{itemize}
  \item Response: a concise and readable advisory opinion tailored to the user's question;
  \item Legal Basis: authoritative statutory references that support the conclusion, including full legal article content.
\end{itemize}

This post-processing step enhances the final output’s legal readability and user trust, ensuring it aligns with real-world standards for legal communication and accountability.

\subsection{Test Set Correction}

This section describes our efforts to ensure rigorous evaluation and high-quality supervision by correcting existing benchmarks and constructing a new dataset for preference-based training. For evaluation, we adopt the widely used LawBench dataset~\citep{fei-etal-2024-lawbench}, which provides question-answer pairs sourced from real-world consultation platforms. However, upon detailed inspection, we identified a substantial number of flawed responses in the test set. These included incorrect legal conclusions, irrelevant or misleading content, and misinterpretations of statutory provisions—issues that severely compromise evaluation reliability.

To address this, we performed a correction of the test set using a hybrid process combining LLM assistance with human expert verification. The LLM was first used to flag potential legal inaccuracies and generate preliminary revision suggestions. These were then reviewed line-by-line by legal professionals to ensure both linguistic clarity and legal correctness. All changes and their justifications are documented in Appendix~\ref{appendix:testset}. We acknowledge the foundational value of LawBench and clarify that our modifications are intended solely to improve the fairness and consistency of evaluation.

\subsection{Construction of dataset and DPO Training}

In addition, we construct a large-scale dataset, \textsc{JurisCQAD}, comprising over 43000 Chinese legal consultation instances derived from real user queries. To train the model’s ability to discern legally sound responses from subtly flawed ones, we generate challenging negative examples using a large language model. To ensure the negative samples are meaningful rather than trivial, we follow a controlled prompt design (see Appendix) that instructs the model to introduce subtle flaws—such as incorrect statutory references, flawed legal reasoning, or conclusions lacking legal support—while maintaining fluency and professionalism. All responses are reviewed by licensed legal professionals to ensure legal accuracy, consistency, and stylistic quality. Each training instance is then organized into a triplet \( (q, y^+, y^-) \), where \( q \) denotes the user query, \( y^+ \) is the expert-approved positive response, and \( y^- \) is the model-generated negative response.

\begin{table}[h]
\centering
\small
\label{tab:juris-stat}
\setlength{\tabcolsep}{4pt} 
\begin{tabular}{l|c|c|c}
\toprule
\textbf{Dataset} & \textbf{QA Pairs} & \textbf{QLength} & \textbf{ALength (Pos./Neg.)} \\
\midrule
\# Train & 39,163 & 15.11 & 263.5 / 193.7 \\
\# Val   & 2,176  & 15.57 & 285.7 / 211.3 \\
\# Test  & 2,176  & 15.22 & 270.8 / 197.5 \\
\bottomrule
\end{tabular}
\caption{Statistics of \textsc{JurisCQAD}.}
\end{table}

The dataset is curated to avoid trivial distractors, such as grammatical errors, misspelled law names, or absurd contradictions. Additionally, we control for potential bias leakage by restricting the model from introducing demographic stereotypes or fabricated statutes. 

Compared to existing legal QA datasets such as LawGPT~\citep{zhou2024lawgptchineselegalknowledgeenhanced}, LawBench~\citep{fei-etal-2024-lawbench}, and JEC-QA~\citep{zhong2020jec}, \textsc{JurisCQAD} offers three key advantages: (1) all queries are sourced post-2021 Civil Code, ensuring up-to-date legal references; (2) it provides higher annotation quality despite being smaller than LawGPT, as shown in table~\ref{tab:finetune_results}; and (3) it uniquely includes expert-verified contrastive supervision with both positive and adversarial responses. See Appendix for a detailed comparison.

We adopt the Direct Preference Optimization (DPO) to train our model on \textsc{JurisCQAD}. Each training instance consists of a user query \( x \), a preferred response \( y^+ \), and a dispreferred response \( y^- \). The objective is to maximize the log-likelihood difference between the two responses:

\begin{equation}
\Delta_{\theta}(x, y^+, y^-) = 
\log \pi_\theta(y^+ \mid x) - \log \pi_\theta(y^- \mid x),
\label{eq:dpo-gap}
\end{equation}
\begin{equation}
\mathcal{P}_{\theta}(x, y^+, y^-) = 
\sigma \left( \beta \cdot \Delta_{\theta}(x, y^+, y^-) \right),
\label{eq:dpo-pref}
\end{equation}
\begin{equation}
\mathcal{L}_{\mathrm{DPO}}(\theta) = 
- \mathbb{E}_{(x, y^+, y^-) \sim \mathcal{D}} 
\left[ \log \mathcal{P}_{\theta}(x, y^+, y^-) \right],
\label{eq:dpo}
\end{equation}

Here, \( \sigma(\cdot) \) denotes the sigmoid function and \( \beta \) is a temperature scaling parameter. This training paradigm enables the model to distinguish legally sound responses from plausible yet flawed ones, improving its ability to generate accurate, well-grounded, and formulated legal advice.

\section{Experiment Details}

\begin{table*}[t]
\centering
\small
\setlength{\tabcolsep}{4.5pt} 
\begin{tabularx}{\textwidth}{l|ccccccccc}
\toprule
\multirow{3}{*}{\textbf{Models}} 
& \multicolumn{3}{c}{\textbf{Rouge (\%)}} 
& \multicolumn{3}{c}{\textbf{Bleu (\%)}} 
& \multirow{3}{*}{\textbf{BertScore (\%)}} 
& \multirow{3}{*}{\textbf{Bleurt (\%)}} 
& \multirow{3}{*}{\textbf{LLM Score}}\\
\cmidrule(lr){2-4} \cmidrule(lr){5-7} 
& \textbf{Rouge-1} & \textbf{Rouge-2} & \textbf{Rouge-L}
& \textbf{Bleu-1} & \textbf{Bleu-2} & \textbf{Bleu-N}\\
\midrule
\multicolumn{9}{c}{\hspace{3em}{\textit{General LLM}}} \\
GPT4o       & 40.24 & 15.27 & 24.50 & \textbf{34.94} & 12.34 & 8.89 & 73.22 & 55.16 & 3.36\\
Qwen3-14B   & 42.55 & 19.24 & 27.27 & 21.93 & 10.25 & 8.10 & \underline{74.64} & \textbf{62.38} & \underline{3.43}\\
Qwen2.5-14B & 40.96 & 16.10 & 23.70 & 34.24 & 13.52 & 9.90 & 73.15 & 56.22 & --\\
\midrule
\multicolumn{9}{c}{\hspace{3em}{\textit{Legal LLM}}} \\
ChatLaw-33B   & 27.78 & 7.40  & 18.15 & 17.84 & 3.81  & 1.46  & 67.77 & 57.54 & 2.30\\ 
Fuzi-mingcha  & 32.31 & 9.92  & 17.41 & 23.13 & 7.07  & 5.19  & 70.47 & 52.46 & 2.78\\
Hanfei        & 30.79 & 10.21 & 18.37 & 13.47 & 4.26  & 2.68  & 69.68 & 58.37 & 2.69\\ 
LawGPT        & 20.52 & 5.22  & 7.26  & 5.15  & 1.01  & 0.49  & 63.44 & 46.94 & 1.52\\
Lawyer-LLaMA  & 30.61 & 9.50  & 18.82 & 24.65 & 6.28  & 3.35  & 69.17 & 58.47 & 2.74\\
LexiLaw       & 31.23 & 9.61  & 18.15 & 14.40 & 4.76  & 3.33  & 70.00 & 58.12 & 2.66\\
Wisdom        & 36.60 & 15.74 & 23.12 & \underline{34.89} & 10.16 & 8.04 & 71.97 & 56.09 & 2.86\\
\midrule
\multicolumn{9}{c}{\hspace{3em}{\textit{Agent / Retrieval Baselines}}} \\
ReAct         & 38.65 & 14.03 & 22.41 & 26.86 & 9.89  & 7.05  & 72.30 & 57.55 & --\\
AutoGen-Flat  & 31.29 & 12.11 & 19.27 & 24.42 & 9.80  & 7.25  & 66.92 & 48.45 & --\\
AutoGen-Tree  & 37.70 & 13.90 & 21.38 & 27.37 & 10.44 & 7.57  & 71.62 & 54.36 & --\\
MMEP          & \underline{42.60} & \underline{20.78} & \underline{27.27} & 30.64 & \underline{14.80} & \underline{12.41} & 73.00 & 56.28 & --\\
LexRAG        & 36.85 & 13.86 & 21.51 & 31.75 & 11.00 & 8.45  & 71.45 & 54.59 & --\\
Parser        & 26.72 & 5.89  & 16.06 & 18.73 & 3.69  & 1.79  & 65.35 & 56.93 & --\\
\midrule
\multicolumn{9}{c}{\hspace{3em}{\textit{Our Method}}} \\
JurisMA   & \textbf{44.68}$^{\dagger}$ & \textbf{23.42}$^{\dagger}$ & \textbf{31.14}$^{\dagger}$ & 32.54 & \textbf{16.18}$^{\dagger}$ & \textbf{14.25}$^{\dagger}$ & \textbf{75.05} & \underline{58.63} & \textbf{3.93} \\
\bottomrule
\end{tabularx}
\caption{
Main results on the revised LawBench. ``$\dagger$'' indicates statistically significant improvement over all baselines under a paired t-test with $p < 0.05$. Bold numbers denote the best performance. Underlined numbers indicate the second-best results.}
\label{tab:main_result}
\end{table*}

\begin{table}[t]
\centering
\small
\setlength{\tabcolsep}{1.5pt} 
\begin{tabular}{lccc}
\toprule
\textbf{Models} & \textbf{Legal Soundness} & \textbf{Reasoning} & \textbf{Completeness} \\
\midrule
GPT4o & 3.54 & 3.38 & 2.88 \\
Lawyer-LLaMA & 2.72 & 2.64 & 2.86 \\
JurisMA & \textbf{4.14} & \textbf{3.58} & \textbf{3.66} \\
\bottomrule
\end{tabular}
\caption{Human evaluation results on three key criteria.}
\label{tab:human_eval}
\end{table}

\subsection{Experimental Setup}

\paragraph{Dataset and Metrics.}
To assess model effectiveness in legal consultation, we evaluate on a revised version of LawBench~\citep{fei-etal-2024-lawbench}, a widely adopted Chinese legal QA benchmark. As discussed in Section 3.5, We corrected Lawbench through a hybrid process of model-assisted detection and expert validation.

We report standard text generation metrics, including ROUGE-1/2/L~\citep{lin2004rouge}, BLEU-1/2/N~\citep{papineni2002bleu}, BERTScore~\citep{zhang2019bertscore}, and BLEURT~\citep{sellam2020bleurt}, computed using official or standard implementations with defaults. Additionally, we include LLM score (GPT-4o) and human validation to assess legal soundness, reasoning, and completeness.

\paragraph{Baselines.}
We compare our method against strong baselines in two groups:

\begin{itemize}
  \item \textbf{General-purpose LLMs}: GPT-4o~\citep{hurst2024gpt}, Qwen3-14B~\citep{yang2025qwen3technicalreport}. These represent state-of-the-art instruction-following LLMs.
  \item \textbf{Legal-domain LLMs}: ChatLaw-33B~\citep{cui2024chatlawmultiagentcollaborativelegal}, Fuzi-Mingcha~\citep{deng-etal-2023-syllogistic}, HanFei~\citep{HanFei}, LawGPT~\citep{zhou2024lawgptchineselegalknowledgeenhanced}, Lawyer-LLaMA~\citep{huang2023lawyerllamatechnicalreport}, LexiLaw~\citep{LexiLaw}, and Wisdom-Interrogatory~\citep{wisdomInterrogatory}. These models are specifically fine-tuned on Chinese legal corpora or tasks.
\end{itemize}

All models are evaluated in a zero-shot setting using a unified prompt schema. Results are averaged over 5 runs with different random seeds to ensure robustness. Details on model configuration, training hyperparameters, and inference setups are provided in Appendix.

\subsection{Main Results}

Table~\ref{tab:main_result} presents our main evaluation results. We summarize the key findings below:

\textbf{(1) Our method achieves the best overall performance across most metrics.} Compared to both general-purpose and legal-specialized LLMs, our approach consistently outperforms all baselines on \textsc{Rouge}, \textsc{Bleu-2/N}, \textsc{BertScore} and \textsc{LLMScore}, and ranks second on \textsc{BLEURT}. This demonstrates the effectiveness of structured legal modeling and agent-based reasoning in enhancing both factual accuracy and legal alignment.

\textbf{(2) Legal-domain LLMs show varied performance depending on their training corpora and alignment strategies.} Models like \textsc{Wisdom} and \textsc{Hanfei} achieve competitive scores, while others lag behind. These differences likely stem from variation in corpus quality, and the degree of alignment with user-facing tasks.

\textbf{(3) Our method excels in both lexical precision and semantic fidelity.} Strong results on \textsc{BERTScore}, \textsc{BLEURT}, and \textsc{LLM Score} (GPT-4o), along with human validation, demonstrate the model’s ability to produce accurate and contextually sound legal responses.

\subsection{Dataset Evaluation}

\begin{figure}[H]
  \centering
  \includegraphics[width=0.5\textwidth]{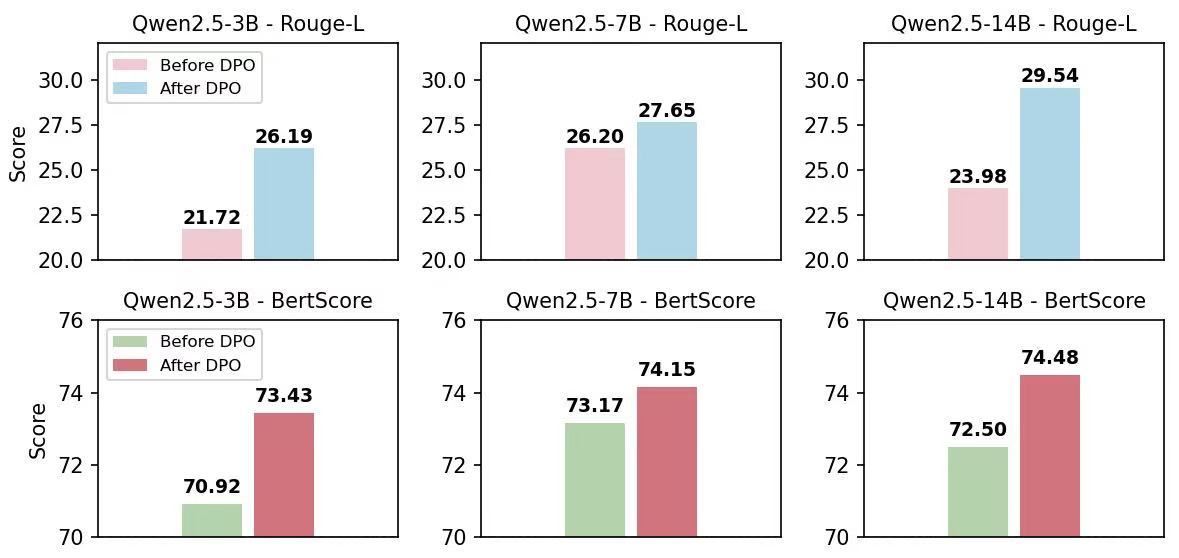}
  \caption{Rouge-L and BertScore comparison before and after DPO across Qwen2.5 models (3B/7B/14B).}
  \label{Figure3_RL}
\end{figure}

To assess dataset effectiveness, we fine-tune Qwen2.5 models (3B/7B/14B) via DPO and report pre-/post-tuning results in Figure~\ref{Figure3_RL}. We also compare our dataset with LawGPT’s corpus by fine-tuning Qwen2.5-3B on both, results in Table~\ref{tab:finetune_results}.

\begin{figure*}[t]
  \centering
  \includegraphics[width=\textwidth]{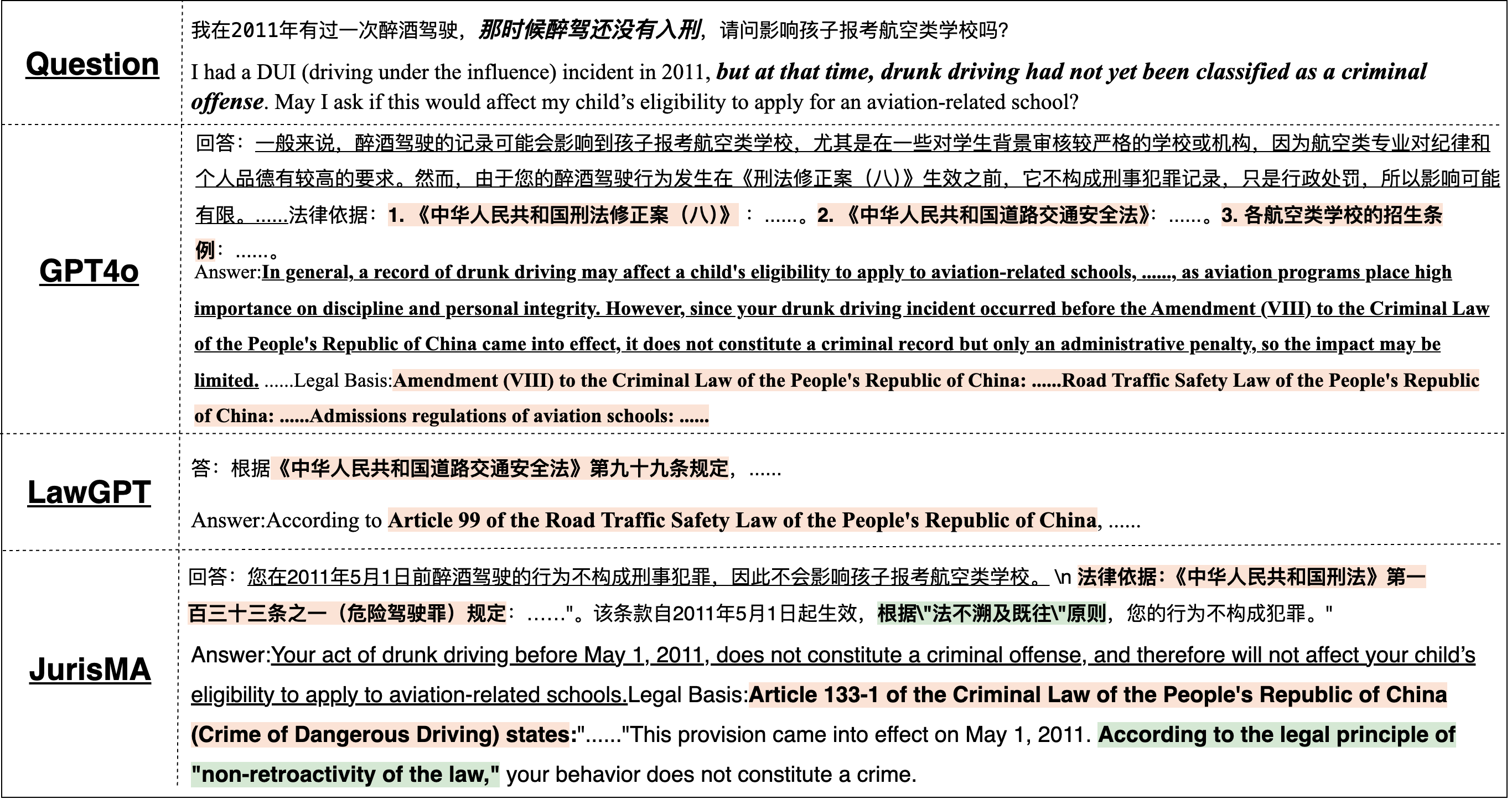}
\caption{
Case study comparing model-generated responses to a time-sensitive legal query. The example illustrates differences in factual interpretation, statutory grounding, and reasoning structure. To aid analysis, we use the following inline annotations: \textbf{bold text} for key legal focus points, \underline{underlined text} for the conclusion of the response, light red highlighting for cited statutes, and light green highlighting for underlying legal principles.
}
  \label{Figure4_Casestudy1}
\end{figure*}

\begin{table}[ht]
\centering
\small
\setlength{\tabcolsep}{2.4pt} 
\begin{tabular}{lcc|cc}
\toprule
\textbf{Model} & \textbf{Rouge-L} & \textbf{Change} & \textbf{BertScore} & \textbf{Change} \\
\midrule
Qwen2.5-3B (Base) & 21.72 & -- & 70.92 & -- \\
+ JurisCQAD & 26.19 & +4.47 & 73.43 & +2.51 \\
+ LawGPT & 20.50 & –1.22 & 68.90 & –2.02 \\
\bottomrule
\end{tabular}
\caption{Performance comparison of Qwen2.5-3B fine-tuned on different training corpora.}
\label{tab:finetune_results}
\end{table}

\indent\textbf{(1) Fine-tuning on \textsc{JurisCQAD} leads to consistent performance gains across all model sizes.} All models show notable improvements in both lexical and semantic metrics. For instance, Qwen2.5-3B improves by +4.47 on \textsc{Rouge-L} and +2.51 on \textsc{BertScore}, while Qwen2.5-14B achieves gains of +5.56 and +1.98.\\
\indent\textbf{(2) \textsc{JurisCQAD} exhibits high annotation quality and practical effectiveness.} As Qwen2.5 models already encode rich Chinese legal knowledge, outdated or noisy data may hinder alignment. In contrast, our curated corpus yields consistent improvements, confirming its superior quality.

\subsection{Ablation Study}
To assess the contribution of each module in our system, we conduct an ablation study by removing key components individually: KG (legal element graph), Manager (decision routing), and Revise process. Table~\ref{tab:ablation} presents results for Qwen2.5-7B and 14B, showing that all modules contribute meaningfully to performance.

\textbf{(1) Revision is critical for semantic quality.} Excluding the revision stage results in the largest \textsc{BertScore} drop (–5.49 for 14B), showing that iterative improves fluency and legal clarity.

\textbf{(2) The manager enhances consistency.} Removing the manager reduces \textsc{Bleu-N} and \textsc{BertScore}, confirming its role in dynamic feedback and controllable generation.

\textbf{(3) The legal graph boosts factual grounding.} Without the legal graph, both \textsc{Rouge-L} and \textsc{BertScore} decline, indicating its importance in encoding legally salient facts.

\begin{table}[t]
\centering
\caption{Ablation study showing performance drops when removing key modules. Results confirm the importance of structured graph input, dynamic task routing, and iterative refinement.}
\small
\setlength{\tabcolsep}{2.8pt} 
\begin{tabularx}{\columnwidth}{l|ccc}
\toprule
\textbf{Models} 
& \textbf{Rouge-L (\%)} 
& \textbf{Bleu-N (\%)} 
& \textbf{BertScore (\%)} \\
\midrule
\multicolumn{4}{l}{\textit{w/o KG}} \\
Qwen2.5:7b       & 21.33 (-6.68) & 7.72 (-3.6) & 71.23 (-2.59) \\
Qwen2.5:14b      & 21.57 (-9.57) & 7.21 (-7.04) & 71.74 (-3.31) \\
\midrule
\multicolumn{4}{l}{\textit{w/o Manager}} \\
Qwen2.5:7b       & 21.41 (-6.6) & 7.81 (-3.51) & 71.62 (-2.2) \\
Qwen2.5:14b      & 21.30 (-9.84) & 6.89 (-7.36) & 71.68 (-3.37) \\
\midrule
\multicolumn{4}{l}{\textit{w/o Revision}} \\
Qwen2.5:7b       & 20.58 (-7.43) & 7.10 (-4.22) & 70.64 (-3.18) \\
Qwen2.5:14b      & 19.96 (-11.18) & 6.10 (-8.15) & 69.56 (-5.49) \\
\midrule
\multicolumn{4}{l}{\textit{Full}} \\
Qwen2.5:7b       & \textbf{28.01} & \textbf{11.32} & \textbf{73.82} \\
Qwen2.5:14b      & \textbf{31.14} & \textbf{14.25} & \textbf{75.05} \\
\bottomrule
\end{tabularx}
\label{tab:ablation}
\end{table}

\subsection{Generalization Evaluation}

To assess the robustness and generalizability of our method beyond the Chinese legal consultation setting, we evaluate JURISMA on three related benchmarks that test cross-lingual, cross-task, and cross-jurisdiction transfer. As detailed in Appendix~\ref{app:generalization}, JURISMA consistently outperforms baselines across all three settings, demonstrating strong generalization capability.

\subsection{Case Study}

To qualitatively assess interpretability and legal reasoning, we present a case study based on a real consultation: whether a man's drunk driving incident in early 2011 would affect his child's eligibility for applying an aviation school. The core legal issue is time-sensitive, as drunk driving was not classified as a criminal offense until May 1, 2011.

Our method delivers the most accurate and concise judgment: the incident occurred before the law came into effect, hence no criminal record, and no eligibility issue. It explicitly cites Article 133-1 of the Criminal Law and invokes the principle of non-retroactivity. In contrast, GPT-4o provides a lengthy, less focused explanation, and \textsc{LawGPT} fails to cite the most directly applicable statute, omitting precise legal grounding.

This example demonstrates our system’s strength in aligning factual analysis with legal authority and reasoning principles, enhancing trustworthiness in high-stakes consultation scenarios

\section{Conclusion}
In this paper, we propose \textsc{JurisMA}, a cognitively inspired multi-agent framework for Legal Consultation Question Answering (Legal CQA). By transforming complex legal queries into structured semantic graphs and coordinating specialized agents via a centralized Manager Agent, our system enables controllable, interpretable, and legally grounded reasoning.

We also introduce \textsc{JurisCQAD}, a dataset of over 43,000 expert-validated Chinese legal consultations. It supports preference-aligned DPO training and enables robust evaluation across diverse scenarios.

Experiments on the corrected LawBench benchmark show that \textsc{JurisMA} outperforms both general-purpose and legal-domain LLMs across multiple lexical and semantic metrics. Ablation studies highlight the importance of structured prompting and dynamic agent routing, while case analyses illustrate improvements in legal consistency and statutory reference quality.

\section*{Limitations}
While \textsc{JurisMA} achieves strong performance, several limitations remain. First, the multi-agent architecture introduces additional latency, which may hinder real-time deployment. Second, although \textsc{JurisCQAD} covers diverse legal scenarios, it may still exhibit bias toward high-frequency consultation topics, limiting generalization to rare or highly specialized cases. Finally, while we incorporate both LLM-based and human evaluations to ensure robustness, broader real-world deployment and continuous user feedback are still needed to assess long-term reliability and practical usability.

\section*{Ethics Statement}
All data used in this study were derived from publicly available Chinese legal
consultation forums.  Personally identifiable information was removed through
automatic anonymization and manual screening before model training.

The expert annotators involved in dataset creation were all members of our research team,
comprising qualified legal professionals. 
One team member holds a Bachelor’s degree in Law and supervised the annotation process. 
All annotators strictly followed the current laws and regulations of the People's Republic of China 
during data review and correction. 
They also received written guidelines covering correctness criteria, bias avoidance, 
and confidentiality requirements. 

During data curation, annotators were instructed not to fabricate legal
provisions and to flag any sensitive or ethically questionable content for
removal. No private, confidential, or client-specific information was retained.
The resulting JurisCQAD dataset is released solely for non-commercial research
under an academic license.

\section*{Acknowledgements}
Mingfei Lu, Yi Zhang, and Mengjia Wu were supported by the Commonwealth Scientific and Industrial Research Organization (CSIRO), Australia, in conjunction with the National Science Foundation (NSF) of the United States, under grant CSIRO-NSF \#2303037.

\bibliography{main}
\appendix

\newpage
\section{More Details for Experimental Setup}
\section*{A.1 JurisCQAD Dataset Details}
\label{sec:JurisCQAD Dataset Details}
The core structure and statistics of the JurisCQAD dataset are summarized in Table 1.
\begin{table}[h]
\centering
\small

\begin{tabular}{p{3.5cm}|p{3.5cm}}
\toprule
\textbf{Property} & \textbf{Description} \\
\midrule
Source & Real-world legal consultation platforms \\
Language & Chinese \\
Size & 43,126 triplets \\
Data Format & (query, positive answer, negative answer) \\
Annotation Method & LLM-assisted generation + expert verification\\
Negative Sample Strategy & LLM-generated distractors with legal/semantic flaws \\
Domains Covered & Contract law, tort liability, family law, labor disputes, etc. \\
Average Query Length & 15.14 tokens \\
Average Positive Answer Length & 264.97 tokens \\
Average Negative Answer Length & 194.79 tokens \\
\bottomrule
\end{tabular}
\caption{Summary of JurisCQAD Dataset}
\end{table}

\begin{table}[t]
\small  
\centering
\renewcommand{\arraystretch}{1.3}
\begin{tabular}{p{1.5cm}|p{2.5cm}|p{2.5cm}}
\hline
\textbf{Comparison Dimension} & \textbf{Legal QA} & \textbf{Legal CQA} \\
\hline
Task Goal & Answer exam questions or legal provisions & Respond to real-world legal concerns from users \\
Task Type & Mostly multiple-choice or extraction tasks & Requires generation of context-relevant legal suggestions \\
\hline
Data Source & Legal exams, statutory texts & Legal forums and Q\&A communities \\
Data Characteristics & Standardized answers, concise questions & Long, complex questions with diverse factual scenarios \\
\hline
Question Structure & Short, standardized text & Long, unstructured, and often informal expressions \\
Legal Context & Involves a single legal provision & Involves multiple statutes and factual elements \\
\hline
Evaluation Metrics & Accuracy, F1 score & BLEU, METEOR, human evaluation (completeness, professionalism, relevance) \\
Answer Diversity & Single correct answer & Multiple plausible answers depending on context \\
\hline
\end{tabular}
\caption{Comparison between Legal QA and Legal Consultation QA (Legal CQA)}
\label{tab:qa_comparison}
\end{table}

\section*{A.2 Implementation Details}
\label{sec:Implementation Details}
We perform DPO fine-tuning on Qwen2.5 models of three different sizes (3B, 7B, 14B). All models are trained with LoRA adapters (rank=8, $\alpha$=16) using the HuggingFace + Deepspeed framework (Stage 2) on up to 2 × A100 80GB GPUs. Gradient accumulation is set to 8, and we use a batch size of 8 per device, for an effective batch size of 128. Mixed precision training is enabled via bf16. All training runs use AdamW with a cosine learning rate schedule, an initial learning rate of $1\times 10^{-5}$, and no warm-up.

The dataset used is JurisCQAD, comprising 43K+ real-world consultation queries with expert-verified (query, positive, negative) triplets. We train for 3 epochs with max sequence length 1024. The DPO $\beta$ is set to 0.1, and the loss is computed using sigmoid preference loss without reward normalization.

All prompts follow the Qwen dialogue template, with system instructions embedded. We do not apply quantization or offloading, and DeepSpeed offload is disabled. Model checkpoints are saved every 100 steps. No external reward model or RLHF phase is used. Evaluation is performed in zero-shot mode using the same prompt template across all models.

In the \textsc{JurisMA}, we employ Qwen2.5-14B-Instruct, fine-tuned on the JurisCQAD dataset, as the underlying model for all agent components. Although newer models such as Qwen3-14B-Instruct have been released with stronger base capabilities, our method—when built upon Qwen2.5—still consistently outperforms Qwen3-14B across all metrics in legal consultation tasks. This highlights the effectiveness of our framework design, independent of backbone improvements. We deliberately avoid using larger models to ensure reproducibility and reduce computational costs, thereby demonstrating that strong performance can be achieved through structural innovation rather than model scaling alone.
All datasets and model baselines used in this study are publicly available under licenses that permit academic use. We ensure that our use is consistent with their intended purpose, strictly limited to research contexts.

Our proposed dataset, \textsc{JurisCQAD}, was constructed from publicly accessible legal consultation forums. All collected samples underwent careful anonymization and manual screening to eliminate personally identifiable information (PII) and potentially offensive content. To ensure ethical integrity, all data was processed solely for non-commercial research use, in line with prevailing data use policies and licensing norms. The dataset will be released for academic purposes only under a research-friendly license.

\section*{A.3 Prompt}
\label{sec:prompt}

The detailed prompts used by each agent in JurisMA are listed in Table~\ref{tab:agent-prompts}.

\begin{table*}[h]
\renewcommand{\arraystretch}{1.3}
\centering
\small
\begin{tabularx}{\linewidth}{p{3.2cm}|X}
\toprule
\textbf{Agent} & \textbf{Prompt Summary} \\
\midrule

\textsc{Negative Example Generation Agent} &
\begin{minipage}[t]{\dimexpr\linewidth-4.2cm}
\small
\vspace{1mm}
Please generate a plausible but flawed response to the following legal consultation question. The response should appear reasonable on the surface but contain either incorrect legal references, flawed logical reasoning, or lack critical statutory support.

\textbf{Requirements:}
\begin{itemize}
    \item The response must be grammatically fluent and professionally styled;
    \item No social biases are allowed (e.g., based on gender, age, ethnicity, or region);
    \item No fabricated or non-existent statute numbers are permitted;
    \item The flaw should be mild—such as subtle legal or factual inaccuracies—not absurd or completely irrelevant;
\end{itemize}

\textbf{Instruction:} Answer the following question. First, provide the flawed response, then specify the legal basis.
\vspace{1mm}
\end{minipage}
\\
\midrule

\textsc{Draft Agent} &
You are a legal consultation draft generation agent. When a user submits a legal-related question, your task is to generate a professional response based on existing legal knowledge.
\\
\midrule

\textsc{Element Agent} &
\begin{minipage}[t]{\dimexpr\linewidth-4.2cm}
\small
\vspace{1mm}
You are a professional legal element extraction expert. Your tasks include:
\begin{itemize}
    \item Extract key case elements from the user's legal consultation;
    \item Identify legal relationships and entities;
    \item Clarify the user's legal demands;
    \item Output a structured element graph in JSON format.
\end{itemize}

\textbf{Element Graph Format:}
\begin{flushleft}
\footnotesize\ttfamily
\{ \\
\quad "entities": [\{ "name": "...", "type": "...", "attributes": \{...\} \}], \\
\quad "events": [\{ "description": "...", "time": "..." \}], \\
\quad "relationships": [\{ "type": "...", "source": ..., "target": ... \}], \\
\quad "user\_claims": ["..."], \\
\quad "key\_facts": ["..."], \\
\quad "legal\_questions": ["..."] \\
\}
\end{flushleft}
\vspace{1mm}
\end{minipage}
\\
\hline
\textsc{Manager Agent} &
\begin{minipage}[t]{\linewidth}
You are the decision-making agent in a multi-agent legal consultation system. Your task is to determine, based on the content of a draft legal response, whether it requires format refinement or legal citation supplementation.\\

\textbf{Decision Criteria:}
\begin{itemize}
    \item If the response is not concise, lacks clear logic, or contains redundancy: \texttt{Call: FormatCheckAgent};
    \item If the response lacks statutory references: \texttt{Call: LawSearchAgent};
    \item If both issues apply: \texttt{Call: FormatCheckAgent} then \texttt{LawSearchAgent};
    \item If the response is acceptable: \texttt{Pass}
\end{itemize}
\end{minipage}
\\
\hline
\textsc{FormatCheck Agent} &
Review the draft for clarity, redundancy, and stylistic issues. Output concrete editing suggestions without changing the legal meaning.
\\
\hline
\textsc{LawSearch Agent} &
Retrieve authoritative legal provisions from Chinese law based on the question and draft response. Output only relevant statute texts.
\\
\hline
\textsc{ContentCheck Agent} &
Rewrite the draft into a fluent, professional legal opinion. Preserve meaning while fusing reasoning and statute into a dual-structured final output.
\\
\bottomrule
\end{tabularx}
\caption{Prompts used by different agents.}
\label{tab:agent-prompts}
\end{table*}

\section{Legal QA \& Legal CQA Comparison}
\label{sec:qa_comparison}
Table~\ref{tab:qa_comparison} provides a comparative overview of traditional Legal QA tasks and the more complex Legal Consultation QA (Legal CQA), highlighting their differences in task objectives, data sources, and evaluation metrics.

\section{Example of Element Graph}
\label{sec:Example of Element Graph}

\begin{table*}[h]
\centering
\footnotesize
\renewcommand{\arraystretch}{1.1}
\begin{tabular}{p{3.2cm}|p{12.5cm}}
\hline
\textbf{Section} & \textbf{Content} \\
\hline

\textbf{Entities} &
\begin{itemize}
  \item \textbf{User (Person)}: Drunk driving record in 2011; the user themself committed the act.
  \item \textbf{Child (Person)}: Child of the user; intends to apply for an aviation school.
  \item \textbf{Drunk Driving (Illegal Act)}: Occurred in 2011; not criminalized at the time (before May 1, 2011).
  \item \textbf{Aviation School (Institution)}: Has specific eligibility requirements for applicants.
\end{itemize} \\
\hline

\textbf{Event} &
\begin{itemize}
  \item \textbf{Description}: The user committed drunk driving in 2011.
  \item \textbf{Time}: 2011
  \item \textbf{Legal Context}: Drunk driving was not yet criminalized before May 1, 2011.
  \item \textbf{Change Effect}: Criminalization started after May 1, 2011, but the user’s act occurred earlier.
\end{itemize} \\
\hline

\textbf{Relationships} &
\begin{itemize}
  \item \textbf{Kinship}: User → Child
  \item \textbf{Application Target}: Child → Aviation School
  \item \textbf{Legal Involvement}: User → Drunk Driving
\end{itemize} \\
\hline

\textbf{User Claims} &
\begin{itemize}
  \item What should I do?
  \item Is it illegal?
  \item Can my child apply to an aviation school?
\end{itemize} \\
\hline

\textbf{Key Facts} &
\begin{itemize}
  \item The user committed drunk driving in 2011.
  \item Drunk driving was not criminalized before May 1, 2011.
  \item The user's child intends to apply for an aviation school.
  \item Aviation schools have specific background requirements.
\end{itemize} \\
\hline

\textbf{Legal Questions} &
\begin{itemize}
  \item Will the drunk driving record affect the child’s application to aviation school?
  \item Was drunk driving a criminal offense in 2011?
  \item What are the background screening standards for aviation school applicants?
\end{itemize} \\
\hline

\end{tabular}
\caption{Example of Element Graph}
\label{tab:element_graph_summary}
\end{table*}

Table~\ref{tab:element_graph_summary} presents each node and its detailed attributes in the element graph extracted from the case study question: “I had a drunk driving incident in 2011. At that time, drunk driving had not yet been criminalized. It was only after May 1st of that year that it became a criminal offense. Will this affect my child's application to an aviation school?”

\section{Test Set Correction Cases}
\label{appendix:testset}
To enhance the reliability and legal validity of evaluation, we manually reviewed and revised a subset of LawBench test cases. Among 500 test queries, 340 were found to contain legal or factual errors and were subsequently corrected. Each correction involved identifying flaws in the original answer, followed by regeneration using expert-reviewed LLMs. Table~\ref{tab:test_corrections} summarizes representative examples and reasons for revision.

\begin{table*}[t]
\centering
\small
\renewcommand{\arraystretch}{1.3}
\setlength{\tabcolsep}{5pt}
\begin{tabularx}{\textwidth}{c|X|X}
\toprule
\textbf{Case ID} & \textbf{Revision Reason} & \textbf{Key Correction} \\
\midrule
1 & 
Original answer failed to distinguish pre-/post-May 1, 2011 legal status of drunk driving and omitted aviation-specific background check regulations. &
Added analysis of non-criminal administrative penalty and cited \textit{Article 133-1 of the Criminal Law} and aviation review guidelines. \\
\midrule
2 &
Original answer discussed unrelated payment default topic and lacked any applicable law to the real estate recovery dispute. &
Rewritten answer clarified legal ownership transfer, invoked \textit{Civil Code} articles on registration, good faith acquisition, inheritance, and statute of limitations. \\
\midrule
3 &
Original answer incorrectly stated that all owners must sign service contracts. It misunderstood the legal effect of contracts signed by the owners’ committee and confused public and private contracting rules. &
Clarified that a legally signed service contract by the owners' committee is binding for all owners under \textit{Civil Code Article 939} and \textit{Property Management Regulations}. \\
\midrule
4 &
Original answer cited outdated or inaccurate medical insurance provisions and failed to reflect local retirement policies. &
Updated answer clarified retirement exemption from further payments, citing \textit{Social Insurance Law} and regional cumulative contribution rules. \\
\midrule
5 &
Original answer misunderstood liability in sublease and construction. Misapplied contract law and omitted tenant’s liability for subtenants’ actions. &
Added correct explanation using \textit{Articles 714, 716, 577 of Civil Code}, showing tenant's liability for third-party damages and breach of duty to maintain the property. \\
\midrule
6 &
Original answer failed to cite core law on execution exemption and missed user's intent to reserve minimum livelihood funds. &
Correction referenced \textit{Civil Procedure Law Article 243} and Supreme Court regulations on exempt property and basic living standards. \\
\midrule
7 &
Original answer did not address user’s question on how to reserve part of pension funds during execution. It also missed the legal basis for such exemption. &
Clarified court must reserve necessary funds during pension account freeze, citing \textit{Civil Procedure Law} and relevant enforcement provisions. \\
\midrule
8 &
Original answer failed to answer whether the drawer could stop check payment. Misquoted irrelevant provisions and missed core check law rules. &
Corrected to include legal conditions under which a drawer may suspend payment, citing \textit{Negotiable Instruments Law}, \textit{Payment and Settlement Measures}, and Supreme Court judicial interpretations. \\
\bottomrule
\end{tabularx}
\caption{Representative Examples of Test Set Corrections}
\label{tab:test_corrections}
\end{table*}

\section{Generalization and Cross-Benchmark Evaluation}
\label{app:generalization}

To further assess the generalization of JurisMA beyond Chinese legal 
consultation, we evaluate it on three related benchmarks that differ from 
our primary training setting along three orthogonal dimensions:
(1) \textbf{language}, using the English \textsc{LegalBench}-RuleQA subset 
to test cross-lingual transfer;
(2) \textbf{task format}, using the LawBench 2--5 Reading Comprehension task 
to test paragraph-level legal understanding; and
(3) \textbf{jurisdiction}, using the KoBLEX benchmark---a Korean legal CQA 
dataset with English translations---to test cross-jurisdiction robustness.
Across all three settings, JurisMA consistently demonstrates strong 
transferability, indicating that the benefits of structured decomposition 
and multi-agent reasoning are not limited to its training distribution.

\subsection{Cross-Lingual Transfer: \textsc{LegalBench}-RuleQA}
\label{app:legalbench}

We first evaluate cross-lingual generalization on the English 
\textsc{LegalBench}-RuleQA subset, which contains factually complex, 
rule-based legal questions. We randomly sample 50 instances and report 
results averaged over 5 random seeds.
As shown in Table~\ref{tab:generalization_result}, JurisMA significantly 
outperforms both general-purpose and legal-domain LLMs across all metrics, 
confirming that our framework transfers effectively to English legal 
reasoning despite being trained exclusively on Chinese consultations.

\subsection{Cross-Task Transfer: LawBench 2--5 Reading Comprehension}
\label{app:lawbench_rc}

We next evaluate cross-task transfer on the LawBench 2--5 Reading 
Comprehension task, which examines paragraph-level legal understanding 
rather than open-ended consultation.
As shown in Table~\ref{tab:lawbench_rc}, JurisMA achieves the second-best 
results despite the substantial difference between this task and legal 
consultation, indicating that our framework retains strong paragraph-level 
legal understanding even when applied to tasks outside its primary 
training objective.

\subsection{Cross-Jurisdiction Transfer: KoBLEX}
\label{app:koblex}

Finally, we test cross-jurisdiction robustness on the KoBLEX benchmark, 
a Korean legal CQA dataset with English translations. This setting 
challenges the model with both a different legal system and a different 
language from its training data.
As shown in Table~\ref{tab:koblex}, JurisMA achieves the best F-1 score 
among all eleven methods, while remaining competitive on Token F-1, 
demonstrating strong effectiveness when transferred to a different 
jurisdiction.

\begin{table*}[t]
\centering
\small
\begin{minipage}[t]{0.48\textwidth}
\centering
\setlength{\tabcolsep}{10pt}
\begin{tabular}{lcc}
\toprule
\textbf{Model} & \textbf{RC-F1} & \textbf{BertScore} \\
\midrule
ChatLaw-33B    & 37.16 & 64.57 \\
Fuzi-mingcha   & \textbf{97.59} & \textbf{84.29} \\
Hanfei         & 38.50 & 64.36 \\
LawGPT         & 2.27  & 51.37 \\
Lawyer-LLaMA   & 34.61 & 63.45 \\
LexiLaw        & 45.39 & 67.01 \\
Wisdom         & 35.56 & 64.13 \\
\midrule
JurisMA  & \underline{46.62} & \underline{70.89} \\
\bottomrule
\end{tabular}
\caption{Results on the LawBench 2--5 Reading Comprehension task.}
\label{tab:lawbench_rc}
\end{minipage}
\hfill
\begin{minipage}[t]{0.48\textwidth}
\centering
\setlength{\tabcolsep}{8pt}
\begin{tabular}{lcc}
\toprule
\textbf{Method} & \textbf{F-1} & \textbf{Token F-1} \\
\midrule
SP                & --    & 30.52 \\
CoT               & --    & 26.37 \\
SP + OR           & 21.50 & 32.18 \\
CoT + OR          & 21.50 & 28.39 \\
Self-Ask          & 9.29  & 16.59 \\
IRCoT             & 20.42 & 31.62 \\
FLARE             & 40.64 & 29.54 \\
ProbTree          & 15.84 & 28.38 \\
BeamAggr          & 14.05 & 16.02 \\
PARSER (original) & \underline{46.24} & \textbf{40.65} \\
\midrule
JurisMA   & \textbf{69.72} & \underline{34.88} \\
\bottomrule
\end{tabular}
\caption{Results on the KoBLEX benchmark.}
\label{tab:koblex}
\end{minipage}
\end{table*}

\begin{table*}[t]
\centering
\small
\setlength{\tabcolsep}{7pt} 
\begin{tabularx}{\textwidth}{l|XXX|XXX|X}
\toprule
\multirow{3}{*}{\textbf{Models}} 
& \multicolumn{3}{c}{\textbf{Rouge (\%)}} 
& \multicolumn{3}{c}{\textbf{Bleu (\%)}} 
& \multirow{3}{*}{\textbf{BertScore (\%)}} \\
\cmidrule(lr){2-4} \cmidrule(lr){5-7} 
& \textbf{Rouge-1} & \textbf{Rouge-2} & \textbf{Rouge-L}
& \textbf{Bleu-1} & \textbf{Bleu-2} & \textbf{Bleu-N}\\
\midrule
\multicolumn{8}{c}{\hspace{3em}{\textit{General LLM}}} \\
GPT4o   & 10.10 & 1.40 & 8.11 & 4.66 & 0.49 & 0.21 & 57.88 \\
\midrule
\multicolumn{8}{c}{\hspace{3em}{\textit{Legal LLM}}} \\
Lawyer-llama  & 8.42 & 0.07 & 7.70 & 3.70 & 0.03 & 0.97 & 57.13 \\
\midrule
\multicolumn{8}{c}{\hspace{3em}{\textit{Our Method}}} \\
JurisMA   & \textbf{20.25}$^{\dagger}$ & \textbf{10.72}$^{\dagger}$ & \textbf{12.81}$^{\dagger}$ & \textbf{16.94}$^{\dagger}$ & \textbf{6.72}$^{\dagger}$ & \textbf{5.41}$^{\dagger}$ & \textbf{70.48}$^{\dagger}$ \\
\bottomrule
\end{tabularx}
\caption{
Generalization results on the \textsc{LegalBench}-RuleQA subset (50 samples, averaged over 5 seeds). “†” indicates statistically significant improvement over all baselines under a paired t-test with \( p < 0.05 \). Bold numbers denote the best performance. Underlined numbers indicate the second-best results.
}
\label{tab:generalization_result}
\end{table*}

\end{document}